\documentclass[runningheads]{llncs}
\usepackage{graphicx}

\usepackage{graphicx}

\usepackage{url}
\usepackage{subcaption}
\usepackage{arydshln}
\usepackage{amssymb}
\usepackage{adjustbox}
\usepackage{soul}
\usepackage{color}
\usepackage{amsmath}

\usepackage{booktabs}
\usepackage{multicol}
\usepackage{multirow}
\usepackage{makecell}
\usepackage{setspace}
\usepackage{bm}
\usepackage{caption}
\usepackage{pifont}

\usepackage[misc]{ifsym}
\begin{document}
\title{A Simple Task-aware Contrastive Local Descriptor Selection Strategy for Few-shot Learning between inter class and intra class}
%
%
\titlerunning{TCDSNet}

\author{Qian Qiao$^{\dag}$
\and Yu Xie$^{\dag}$ 
\and Shaoyao Huang
\and Fanzhang Li \thanks{Corresponding Author. \dag Equal contribution.}}

%
\authorrunning{Q. Qiao et al.}
\institute{School of Computer Science and Technology, Soochow University, Suzhou, 215006, China\\
\email{\{qqiao, 20224227022\}@stu.suda.edu.cn}\\
\email{xieyu20001003@gmail.com}\\
\email{lfzh@suda.edu.cn}}
\maketitle              
\begin{abstract}
Few-shot image classification aims to classify novel classes with few labeled samples.
Recent research indicates that deep local descriptors have better representational capabilities. These studies recognize the impact of background noise on classification performance. They typically filter query descriptors using all local descriptors in the support classes or engage in bidirectional selection between local descriptors in support and query sets. However, they ignore the fact that background features may be useful for the classification performance of specific tasks.
This paper proposes a novel task-aware contrastive local descriptor selection network (TCDSNet). First, we calculate the contrastive discriminative score for each local descriptor in the support class, and select discriminative local descriptors to form a support descriptor subset. Finally, we leverage support descriptor subsets to adaptively select discriminative query descriptors for specific tasks. 
Extensive experiments demonstrate that our method outperforms state-of-the-art methods on both general and fine-grained datasets.
\keywords{few-shot learning\and task-aware\and local descriptor\and image classification}
\end{abstract}

\section{Introduction}
\label{sec:intro}
The purpose of few-shot learning is to enable models to adapt quickly to new tasks with only a small number of training samples in scenarios where data is scarce. Generally, these methods can be divided into three groups: optimization-based methods \cite{finn2017model,antoniou2018train,jiang2022continual}, metric-based methods \cite{snell2017prototypical,sung2018learning,jiang2022lie}, and data augmentation-based methods \cite{antoniou2017data,schwartz2018delta,xian2019f,huang2023diffusion,he2024freestyle}.

This work is based on a few-shot learning method using local descriptors, falling within the realm of metric learning. Features based on local descriptors exhibit superior representational capabilities compared to image-level features. In previous works, \cite{li2019revisiting} proposed DN4, which directly utilizes all query descriptors. It selects k support descriptors directly for each query local descriptor through k-nearest neighbors (k-NN) and approximates the relationship between query samples and support classes using cosine similarity distances. Based on DN4, \cite{liu2022dmn4} introduced DMN4, which believes that not all query descriptors are task-relevant and contain significant background noise. DMN4 establishes mutual nearest neighbor (MNN) relationships, explicitly selecting query descriptors most relevant to each task, thereby avoiding the impact of background noise on classification performance. Similarly, based on DN4, \cite{dong2021learning} and \cite{yan2023few} proposed ATL-Net and TADNet, respectively. Both methods measure the relationships between each query local descriptor and all support classes, adaptively selecting discriminative query descriptors for classification. \cite{qiao2024talds} introduced TALDS-Net, recognizing background noise in query descriptors. It first adaptively selects optimal descriptor subsets composed of support class local descriptors and then adaptively chooses query descriptors for classification from the optimal descriptor subset. However, all these methods aim to eliminate background noise to prevent its influence on the feature representation of local descriptors, either by filtering query descriptors through all support class local descriptors or by bidirectionally selecting between support class local descriptors and query descriptors. Their goal is to remove background noise.
We observe that from a human cognitive perspective, for example, considering an image of a dog and an image of a dolphin, not only do the target features differ significantly, but the background features also exhibit substantial distinctions (\emph{e.g.}, dolphin backgrounds are unlikely to be grassy, whereas dog backgrounds might include grass). In such cases, background features can contribute to classification. Conversely, for two images both belonging to the dolphin category, the differences in background features might not be as pronounced, and in this scenario, background features can be considered as noise. For instance, when dealing with unfamiliar images, if the background is an ocean, it can help narrow down the classification to objects commonly found only in the ocean. This aids in identifying the target category among familiar images. Thus, background features within the same category might positively impact classification performance. Furthermore, background features between different categories might also contribute to enhancing classification performance. Determining discriminative local descriptors for methods based on local descriptors is a challenging task. Moreover, it is essential to judiciously retain or discard background noise in the process.

In response to this challenge, a straightforward solution is to select local descriptors in the support class to form a support descriptor subset, and then use the support descriptor subset to select query descriptors. Experimental results have also demonstrated the effectiveness of this simple method.

The above-described method is our proposed Task-Aware Contrastive Discriminative Local Descriptor Selection Network (TCDSNet). Specifically, we first select local descriptors from the support class. For each support descriptor, we calculate the sum of its similarities with the remaining support descriptors in the same category as the intra-class similarity score. Next, we compute the sum of its similarities with support descriptors from other categories as the inter-class similarity score. A high intra-class similarity score indicates that the support descriptor has strong representational capabilities for that class, while a low inter-class similarity score suggests that the support descriptor has high discriminative power across other classes. We calculate the discriminative score by dividing the intra-class similarity score by the inter-class similarity score, which we term the contrastive discriminative score. Then we select $K$ support descriptors in descending order of their discriminative scores. Finally, we utilize the selected support descriptors to choose query descriptors. For selecting discriminative query descriptors, we employ a simple learnable module to adaptively predict a threshold. Using the learned threshold and a score map, we select the most discriminative descriptors for final classification. This approach enhances the model's classification and generalization capabilities.

In summary, our main contributions are three folds:
\begin{itemize}
\item We propose a novel method that calculates discriminative scores ($\mathcal{CDS}$) for local descriptors in the support class. This enhances the model's adaptability to different tasks and strengthens the performance of local descriptors in few-shot learning tasks.

\item We propose a novel Task-Aware Contrastive Discriminative Local Descriptor Selection Network (TCDSNet) that not only selects a subset of support descriptors based on discriminative scores but also incorporates a learnable module for adaptively choosing the discriminative query descriptors. 

\item Extensive experimental results demonstrate that TCDSNet outperforms state-of-the-art methods on multiple general and fine-grained datasets.
\end{itemize}

\section{METHOD}
\label{sec:method}
\begin{figure*}[!t]
\setlength{\abovecaptionskip}{0.5cm}
\setlength{\belowcaptionskip}{-0.5cm}
\centerline{\includegraphics[width=1.0\linewidth]{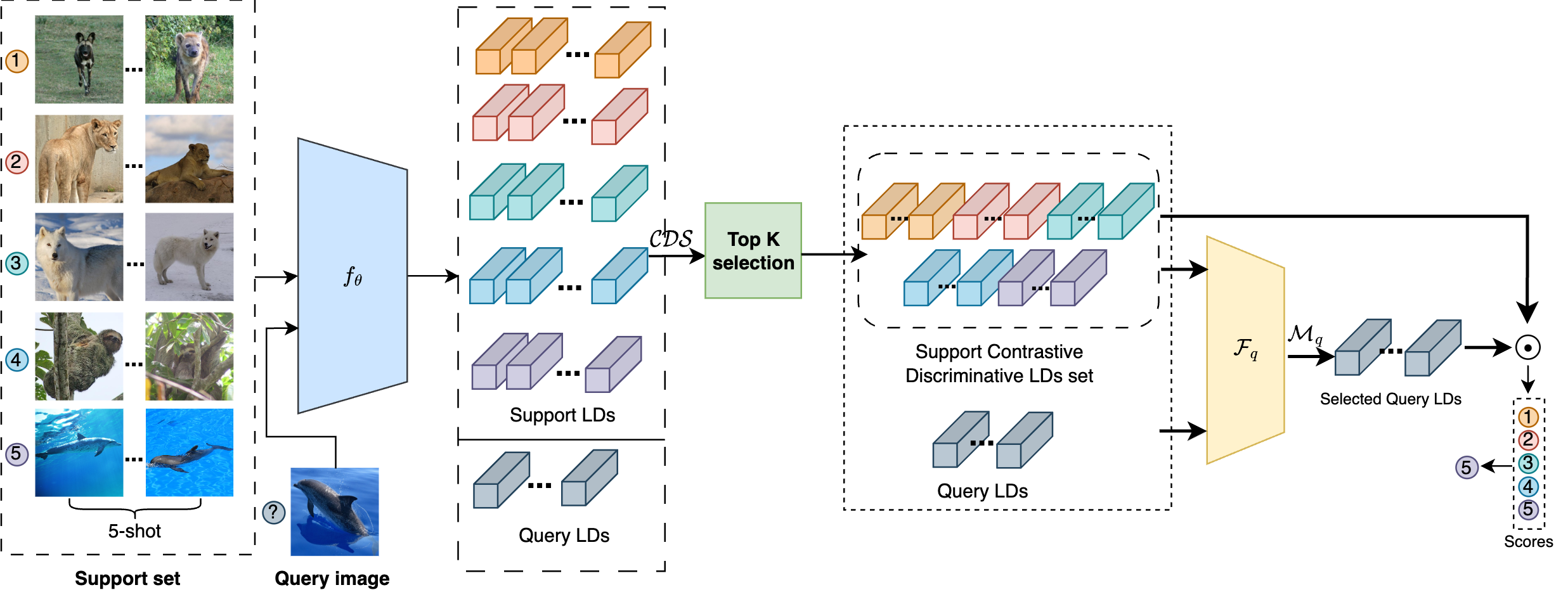}}
\caption{The overall architecture of the proposed method under a 5-way 5-shot setting. The model primarily consists of three components: a feature extraction module $f_{\theta}$ for extracting features, a model for selecting $K$ discriminative LDs, and $\mathcal{F}_q$ for adaptively selecting query LDs.}
\label{fig:1}       
\end{figure*}

Fig. \ref{fig:1} shows an overview of the proposed method. 

\subsection{Problem Definition}
In this paper, we follow the same setting as previous methods\cite{li2019revisiting,dong2021learning,yan2023few,qiao2024talds}. Given a support set $S$, a query set $Q$, and an auxiliary set $A$, where the label space of the auxiliary set $A$ is disjoint from $S$ and is used to learn transferable knowledge. The support set $S$ contains $C$ classes, each with $K$ labeled samples, while the samples in the query set $Q$ are unlabeled and share the same label space as $S$. We are given a support set consisting of $n$ classes, each with $k$ samples, and a query image, and the task is to classify the query image into one of the $n$ support classes. This constitutes the n-way k-shot few-shot classification problem. Under this setting, we introduce a meta-training mechanism\cite{vinyals2016matching} called the episodic training mechanism. We randomly sample from the auxiliary set $A$ to construct an n-way k-shot task. Each task consists of a support set $A_S$ and a query set $A_Q$. During the training phase, we construct tens of thousands of tasks to learn transferable knowledge.

\subsection{Image Representation Based on Local Descriptors}
\label{subsec:irbLDs}
We obtain a three-dimensional feature representation $f_{\theta}(X)\in \mathbb{R}^{h\times w\times d}$ for the image $X$ through the embedding module $f_{\theta}(\cdot)$, which is considered as a set of $d$-dimensional local descriptors (LDs):
\begin{equation}
\begin{split}
    f_{\theta}(X)=[l_1,l_2,\cdots,l_m]\in \mathbb{R}^{m \times d}
\end{split}
\end{equation}
Where $l_i$ denotes the $i$-th deep local descriptor (LD). Similar to other descriptor-based Methods \cite{li2019revisiting,dong2021learning,liu2022dmn4,yan2023few,qiao2024talds}, we consider it as a set of $m$ $d$-dimensional descriptors, and $m = h \times w$.
\par
In each episode, each support class has $k$ images. We denote the descriptor set from category $c$ as $\mathcal{L}_c^S$, where there are $n$ classes in total, and represent the descriptor representation for each query image as $l^q$. When using shallower embedding modules (\emph{e.g.}, Conv-4), each support category is represented in its original form. When using deeper embedding modules (\emph{e.g.}, ResNet-12), each support category is represented by the empirical mean of its support descriptors.

\begin{table*}
\caption{The classification accuracies on the miniImageNet and tieredImageNet datasets in the 5-way 1-shot and 5-shot settings using Conv-4 and ResNet-12 as backbones with $95\%$ confidence interval. All the results of comparative methods are from the exiting literature ('-' not reported). The methods below "hline" are LDs-based methods.}
\label{tab1}
\resizebox{\linewidth}{!}{
\begin{tabular}{lcccccccc}
\toprule
\multicolumn{1}{c}{\multirow{3}{*}{\textbf{Method}}} & \multicolumn{4}{c}{\textbf{Conv-4}}                        & \multicolumn{4}{c}{\textbf{ResNet-12}} \\
\multicolumn{1}{c}{} &
  \multicolumn{2}{c}{miniImageNet} &
  \multicolumn{2}{c}{tieredImageNet} &
  \multicolumn{2}{c}{miniImageNet} &
  \multicolumn{2}{c}{tieredImageNet} \\ 
  \cmidrule(r){2-3} \cmidrule(r){4-5} \cmidrule(r){6-7} \cmidrule(r){8-9}
\midrule
MatchingNet\cite{vinyals2016matching}   & 43.56 $\pm$ 0.84 & 55.31$\pm$0.73 & - & - & 63.08$\pm$0.20 & 75.99$\pm$0.15 & 68.50$\pm$0.92 & 80.60$\pm$0.71 \\
ProtoNet\cite{snell2017prototypical}     & 51.20$\pm$0.26 & 68.94$\pm$0.78 & 53.45$\pm$0.15 & 72.32$\pm$0.57 & 62.33$\pm$0.12 & 80.88$\pm$0.41 & 68.40$\pm$0.14 & 84.06$\pm$0.26 \\
RelationNet\cite{sung2018learning}  & 50.44$\pm$0.82 & 65.32$\pm$0.70 & 54.48$\pm$0.93 & 71.31$\pm$0.78 & 60.97  & 75.12   & 64.71 & 78.41 \\
FRN\cite{wertheimer2021few}  & 54.87  & 71.56 & 55.54 & 74.68 & 66.45$\pm$0.19 & 82.83$\pm$0.13 & 72.06$\pm$0.22 & 86.89$\pm$0.14 \\
Meta-OLE\cite{wang2023meta}  & 56.82$\pm$0.84 &73.87$\pm$0.67& 58.82$\pm$0.88 & 75.85$\pm$0.87 & 67.04$\pm$0.72 & 82.23$\pm$0.67 & 68.82$\pm$0.71 & 85.51$\pm$0.59\\
Approximate GAP\cite{kang2023meta}&53.52$\pm$0.88 &70.75$\pm$0.67&57.47$\pm$0.99& 71.66$\pm$0.76&-&-&-&-\\
GAP\cite{kang2023meta}&54.86$\pm$0.85 &71.55$\pm$0.61&58.56$\pm$0.93& 72.82 $\pm$0.77&-&-&-&-\\
\midrule
DeepEMD\cite{zhang2020deepemd}   & 51.72$\pm$0.20 & 65.10$\pm$0.39 & 51.22$\pm$0.14 & 65.81$\pm$0.68 & 65.91$\pm$0.82 & 82.41$\pm$0.56 & 71.16$\pm$0.87 & 86.03$\pm$0.58 \\
DN4\cite{li2019revisiting}   & 51.24$\pm$0.74 & 71.02$\pm$0.64 & 52.89$\pm$0.23 & 73.36$\pm$0.73 & 65.35 & 81.10 & 69.60 & 83.41 \\
DMN4\cite{liu2022dmn4} & 55.77  & 74.22 & 56.99 & 74.13 & 66.58  & 83.52  & 72.10 & 85.72  \\
ATL-Net\cite{dong2021learning}   & 54.30$\pm$0.76 & 73.22$\pm$0.63 & - & -  & -  & -  & - & -  \\
TADNet\cite{yan2023few} & 56.14 $\pm$ 0.20 & 74.68$\pm$0.15 & 57.88$\pm$0.21 & 75.98$\pm$0.17 & 67.26$\pm$0.20 & 84.23$\pm$0.13 & 71.29$\pm$ 0.22 & 86.46$\pm$0.15\\
\midrule
\textbf{TCDSNet(ours)} &
  \multicolumn{1}{l}{\textbf{57.14$\pm$0.22}} &
  \multicolumn{1}{l}{\textbf{75.89$\pm$0.35}} &
  \multicolumn{1}{l}{\textbf{58.67$\pm$0.61}} &
  \multicolumn{1}{l}{\textbf{76.06$\pm$0.33}} &
  \multicolumn{1}{l}{\textbf{68.53$\pm$0.19}} &
  \multicolumn{1}{l}{\textbf{85.12$\pm$0.42}} &
  \multicolumn{1}{l}{\textbf{72.43$\pm$0.72}} &
  \multicolumn{1}{l}{\textbf{87.35$\pm$0.55}} \\
  \bottomrule
\end{tabular}}
\end{table*}

\subsection{Contrastive Discriminative Scores for Support Local Descriptors Selection}

As mentioned above, $X_S$ represents an image in the support class, fed into the embedding module $f_{\theta}$ to obtain local descriptors $\mathcal{L}^S=f_{\theta}(X_S)\in \mathbb{R}^{m\times d}$, where $m=h\times w$. Here, $l^s$ denotes a supporting local descriptor in $\mathcal{L}^S$, $\hat{\mathcal{L}}^S$ represents the set of remaining support descriptors in $\mathcal{L}_S$ excluding the current $l^s$, and $\bar{\mathcal{L}}^S$ represents the set of local descriptors from the remaining support classes. Thus, we obtain $m$ $d$-dimensional local descriptors (LDs) for an image in the support class. Under the n-way k-shot setting, there are a total of $nkm$ $d$-dimensional support LDs.
Previous methods \cite{qiao2024talds} only considered the average similarity between each LD and the remaining LDs within the same class as the discriminative score. However, our goal is not only to maintain discriminative relationships within the same class but also across other classes. For each $l^s$, we calculate its average similarity with all other LDs within the same support class, referred to as intra-class similarity, and then calculate its average similarity with LDs from the remaining support classes, referred to as inter-class similarity. We seek support LDs with high intra-class similarity and low inter-class similarity. High intra-class similarity indicates strong representational capabilities of the support LD for its corresponding class, while low inter-class similarity signifies poorer representational capabilities of the support LD for other classes. Support LDs exhibiting these characteristics suggest discriminative capabilities, potentially incorporating discriminative background features to enhance classification results. Therefore, the calculations for intra-class and inter-class similarities are as follows:
\begin{equation}
    \begin{split}
        \text{SIM}_{intra} = \frac{1}{m-1} \sum_{\hat{l}^s\in \hat{\mathcal{L}}^S} \text{cos}(l^s, \hat{l}^s),\\
        \text{SIM}_{inter} = \frac{1}{(n-1)m}\sum_{\bar{l}^s\in \bar{\mathcal{L}}^S} \text{cos}(l^s, \bar{l}^s)\\
    \end{split}
\end{equation}
Where $\hat{\mathcal{L}}^S$ represents the set of remaining support descriptors in $\mathcal{L}_S$ excluding the current $l^s$ (in the case of 1-shot, it corresponds to the remaining local descriptors of the current image). $\bar{\mathcal{L}}^S$ denotes the set of local descriptors from the remaining support classes, $\text{SIM}_{intra}$ denotes the intra-class similarity score, and $\text{SIM}_{inter}$ denotes the inter-class similarity score. Furthermore, we normalize these two similarity scores and subsequently calculate their discriminative scores:
\begin{equation}
    \begin{split}
        \mathcal{D}_{intra} = \text{softmax}(\text{SIM}_{intra})\\
        \mathcal{D}_{inter} = \text{softmax}(\text{SIM}_{inter})
    \end{split}
\end{equation}
Where $\mathcal{D}_{intra}$ denotes the discriminative score of the local descriptor $l^s$ within its own class, and $\mathcal{D}_{inter}$ represents its discriminative score across classes.
\par
\begin{figure}[!t]
\setlength{\abovecaptionskip}{0.5cm}
\setlength{\belowcaptionskip}{-0.5cm}
\vspace{1em}
\begin{spacing}{0.6}
\centering
\includegraphics[width=1\linewidth, height=0.50\linewidth]{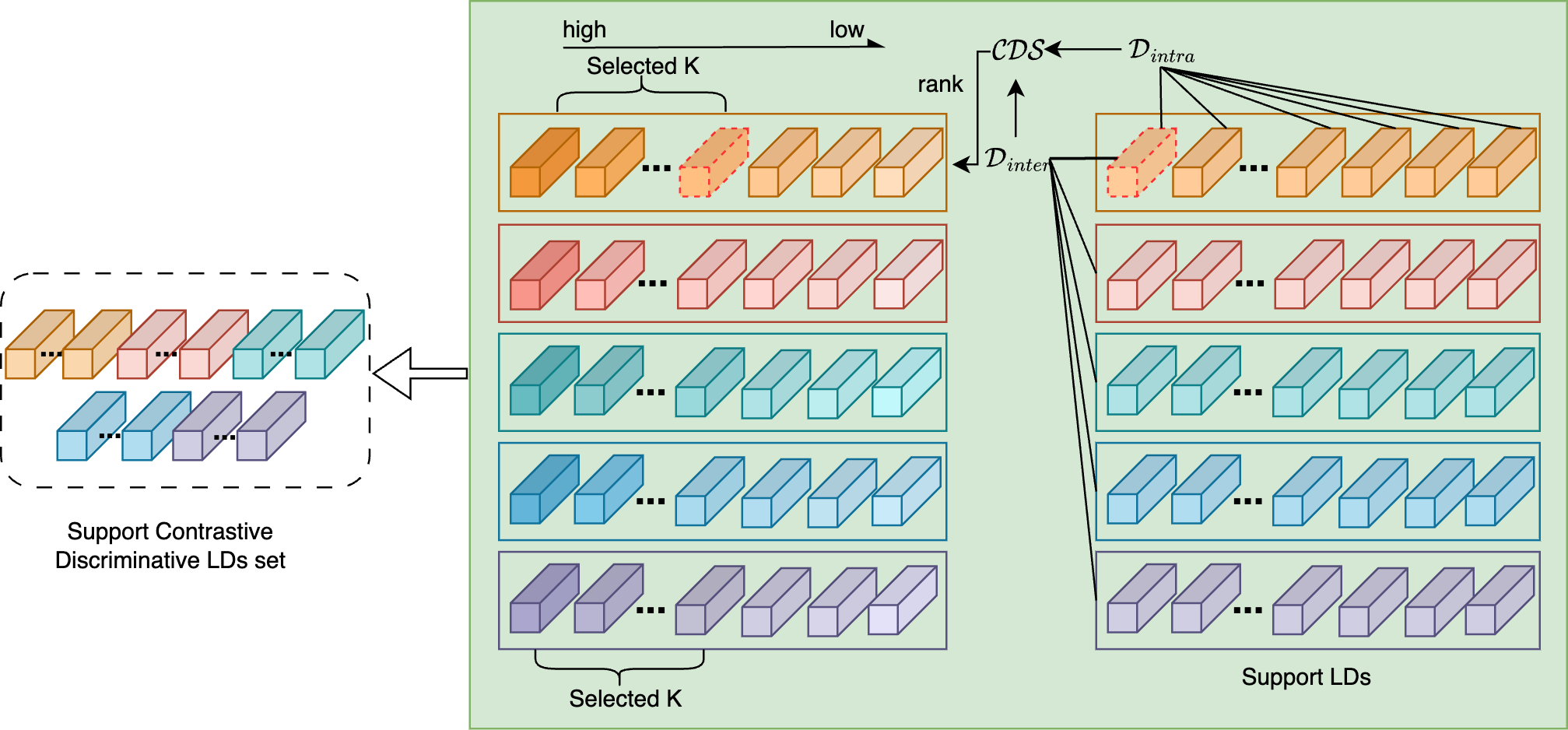}
\vspace{-1em}\caption{Selecting $K$ LDs to form a discriminative support descriptor set is achieved by computing the $CDS$ for each LD in each support class.}
\label{fig:2}
\end{spacing}
\vspace{0.5em}
\end{figure}

Based on the above results, we can calculate the two discriminative scores $\mathcal{D}_{intra}$ and $\mathcal{D}_{inter}$ for each support descriptor using a comparative approach. Subsequently, an optimized Contrastive Discriminative Score ($\mathcal{CDS}$) can be computed:
\begin{equation}
    \begin{split}
        \mathcal{CDS} = \sigma(\frac{\mathcal{D}_{intra}}{\mathcal{D}_{inter}})
    \end{split}
\end{equation}
We can observe that $\mathcal{CDS}$ aligns well with our initial idea, indicating that the current local descriptor exhibits high similarity with other local descriptors within the same class and low similarity with local descriptors from other classes, where $\sigma$ is a sigmoid function. Furthermore, in Fig. \ref{fig:2}, based on the descending order of $\mathcal{CDS}$, we select the top $K$ support descriptors with the contrastive discriminative scores for each class, forming a discriminative support descriptor set:
\begin{equation}
    \begin{split}
        \mathcal{L}_c^{\mathcal{CDS}} = \text{Top K}_{l_i} (\mathcal{CDS})
    \end{split}
\end{equation}
The value of $K$ will be discussed in Ablation Studies \ref{ablation_K}. We will form a set $\mathcal{L}_{\mathcal{CDS}}$ with the support descriptors selected after screening.

\subsection{Query Local Descriptors Selection}
Given a query image $X_q$ embedded as $\mathcal{L}^Q = f_{\theta}(X_q)\in \mathbb{R}^{m\times d}$. $l^q_i$ denotes a query descriptor in $\mathcal{L}^Q$. Previous works\cite{yan2023few,qiao2024talds} employed $k$-NN to select $k$ support descriptors from each support class. However, we have observed that, after computing the discriminative support descriptor set $\mathcal{L}^\mathcal{CDS}$, it is not necessary to use $k$-NN for selecting $k$ support LDs from $\mathcal{L}^\mathcal{CDS}$. We directly compute the sum of similarities between each query descriptor $l_i^{q}$ and the discriminative support descriptor set $\mathcal{L}_c^{\mathcal{CDS}}$ for each support class $c$:
\begin{equation}
    \begin{split}
        \text{SIM}_c^{l_i^{q}}=\sum_{l_c\in \mathcal{L}_c^{\mathcal{CDS}}} \text{cos}(l_i^q,l_c)
    \end{split}
\end{equation}
Where $c \in \{1, 2, \ldots, k\} $ denotes a support class, and $l_c$ is one discriminative support LD from the discriminative support descriptor set of support class $c$. Similarly, the discriminative score for each query descriptor $l_i^{q}$ is calculated as:
\begin{equation}
    \begin{split}
        \mathcal{D}^{{l_i^{q}}}= \max_c \left(\frac{\text{SIM}_c^{l_i^{q}}}{\sum_{c=1}^n \text{SIM}_c^{l_i^{q}}}\right)
    \end{split}
\end{equation}

Previous works \cite{li2019revisiting,liu2022dmn4,dong2021learning,yan2023few,qiao2024talds} have employed methods that directly select query descriptors by using a fixed threshold $\mathcal{V}$ and the top-$k$ query descriptors with the highest similarity. However, both of these methods suffer from poor generalization, as they may overlook some discriminative LDs. 
Thus, inspired by \cite{dong2021learning,yan2023few,qiao2024talds,huang2024contrastive}, we employ a network $\mathcal{F}_q$ consisting of two fully connected layers as an MLP to adaptively predict the threshold $\mathcal{V}^{l_i^q}$ for each query descriptor. Finally, we utilize the predicted threshold $\mathcal{V}$ to learn a query descriptor weights map $\mathcal{M}_q$.
We feed the discriminative support descriptor set $\mathcal{L}_c^{\mathcal{CDS}}$ and the query descriptor $l_i^q$ into $\mathcal{F}_q$, ultimately predicting the threshold $\mathcal{V}^{l_i^q}$:
\begin{equation}
    \begin{split}
        \mathcal{V}^{l_i^q}=\sigma(\mathcal{F}_q(l_i^q, \mathcal{L}_c^{\mathcal{CDS}}))
    \end{split}
\end{equation}
Where $i\in\{1, 2, \ldots, m\}$ denotes a query LD, and $c\in\{1, 2, \ldots, n\}$ denotes a support class. The final calculation for the query descriptor weights map $\mathcal{M}_q$ is as follows:
\begin{equation}
    \begin{split}
        &\mathcal{M}_q = {1}/{(1+\exp^{-\lambda(\mathcal{D}^{{l_i^{q}}}-\mathcal{V}^{l_i^q})})}
    \end{split}
\end{equation}
Where, when $\lambda$ is sufficiently large and $\mathcal{D}^{{l_i^{q}}} >\mathcal{V}^{l_i^q}$, the values of $\mathcal{M}_q$ approximates $1$. Conversely, the values of $\mathcal{M}_q$ approximates $0$. 
\par
Therefore, we can utilize $\mathcal{M}_q$ to select query LDs. The calculation for the similarity scores between each query image $X_q$ and each support class $c$ is as follows:
\begin{equation}
    \begin{split}
        \text{Score}(X_q,c)=\sum_{l^q_i\in \mathcal{L}_c^Q} \mathcal{V}^{l^q_i} \mathcal{M}_q
    \end{split}
\end{equation}
The cross-entropy loss is used to meta-train the network:
\begin{equation}
    \begin{split}
        &p_\phi\left(y=c \mid X_q\right)=\frac{\exp \left(\text{score}(X_q,c)\right)}{\sum_{c^{\prime}=1}^n \exp \left(\text{score}(X_q,c^{\prime})\right)}
    \end{split}
\end{equation}
\begin{equation}
    \begin{split}
        &\mathcal{J}({\phi})=-\frac{1}{|A_Q|}\sum_{X_q\in A_Q}\sum_{c=1}^n y\log p_\phi\left(y=c \mid X_q\right)\\ 
    \end{split}
\end{equation}

\section{EXPERIMENTS}
\label{sec:experiments}
In this section, we validate the effectiveness of our proposed method on several few-shot benchmark datasets and compare it with other state-of-the-art LDs-based methods. Additionally, we compare our method with small-sample methods using different parameter settings. Furthermore, we conduct ablation experiments to further analyze and validate the effectiveness of our proposed method.
\subsection{Datasets}
\textbf{miniImageNet}\cite{vinyals2016matching} is a subset of ImageNet \cite{deng2009imagenet}. It is divided into a training set with $64$ classes, a validation set with $16$ classes, and a test set with $20$ classes. Each class consists of $600$ image samples, each of size $84 \times 84$ pixels. 
\par
\textbf{tieredImageNet}\cite{ren2018meta} is another subset of ImageNet. It comprises $608$ classes, with each class containing $1281$ images. These $608$ classes are divided into $351$ for training, $97$ for validation, and $160$ for testing. 
\par
\textbf{CUB-200}\cite{wah2011caltech} is a fine-grained dataset that consists of $11788$ bird images, encompassing $200$ different bird species. We partition it into $100$ classes for training, $50$ classes for validation, and $50$ classes for testing. For fine-grained datasets, we resized the images in them to the same size as miniImageNet, which is $84\times 84$ pixels.

\begin{table}[!t]
\caption{
The classification accuracies on the CUB-200 dataset in the 5-way 1-shot and 5-shot settings using Conv-4 and ResNet-12 as backbones, The confidence intervals of our method are all below 0.20. }
\label{tab2}
\centering
\begin{tabular}{lcccc}
\toprule
\multicolumn{1}{c}{\multirow{2}{*}{\textbf{Method}}} & \multicolumn{2}{c}{\textbf{Conv-4}}    & \multicolumn{2}{c}{\textbf{ResNet-12}}       \\ \cmidrule(r){2-3} \cmidrule(r){4-5}
\multicolumn{1}{c}{} & 1-shot & 5-shot & 1-shot & 5-shot \\ 
\midrule
ProtoNet\cite{snell2017prototypical}  & 63.73  & 81.50  & 66.09  & 82.50  \\
DSN\cite{simon2020adaptive}  & 66.01  & 85.41  & 80.80  & 91.19  \\
FRN\cite{wertheimer2021few}   & 73.48  & 88.43  & 83.16  & 92.59  \\
Meta-OLE\cite{wang2023meta} & 71.32  & 86.11 & - & -\\
Approximate GAP\cite{kang2023meta}&43.77& 62.92  &-&-\\
GAP\cite{kang2023meta} & 44.74 & 64.88& - & -\\
DeepEMD\cite{zhang2020deepemd}  & -  & -  & 77.14  & 88.98  \\
DN4\cite{li2019revisiting}   & 73.42  & 90.38  & -   & -   \\
DMN4\cite{liu2022dmn4}  & 78.36  & 92.16  & -  & - \\ 
TADNet\cite{yan2023few} & 82.47 & 93.36 & 87.62 & 94.80\\
\midrule
\textbf{TCDSNet(ours)}  & \multicolumn{1}{c}{\textbf{82.73}} & \multicolumn{1}{c}{\textbf{95.04}} & \multicolumn{1}{c}{\textbf{88.71}} & \multicolumn{1}{c}{\textbf{95.82}} \\ 
\bottomrule
\end{tabular}
\end{table}

\begin{table}[!t]
\caption{Ablation study on miniImageNet and CUB-200 datasets for the influence of Top $K$ in support LDs selection.}
\label{tab_topk}
\centering
\begin{tabular}{lcc|lcc}
\toprule
\multicolumn{3}{c|}{\textbf{Conv-4}}            & \multicolumn{3}{c}{\textbf{ResNet-12}}          \\ 
\midrule
\textbf{K} & \multicolumn{1}{l}{miniImagenet} & \multicolumn{1}{l|}{CUB-200} & \textbf{K} & minImagenet & \multicolumn{1}{l}{CUB-200} \\ 
\midrule
1\%  & 74.94          & 90.11          & 3\%  & 83.92          & 89.21          \\
2\%  & \textbf{75.89} & 90.23          & 5\%  & \textbf{85.12} & 89.25          \\
5\%  & 74.23          & 92.37          & 10\% & 84.39          & 92.33          \\
10\% & 72.02          & \textbf{95.04} & 25\% & 84.41          & \textbf{95.82}\\
30\% & 71.11          & 94.57          & 30\% & 83.83          & 94.29          \\
\bottomrule
\end{tabular}
\end{table}

\subsection{Implementation Details}
\textbf{Model architecture.} We use Conv-4 and ResNet-12 as feature extraction networks $f_{\theta}$, similar to previous work \cite{li2019revisiting,dong2021learning,yan2023few,qiao2024talds}. Conv-4 consists of 4 convolutional blocks, each containing a convolutional layer, batch normalization layer, and Leaky ReLU layer. ResNet-12 is composed of 4 residual blocks, with each block consisting of 3 convolutional layers with $3\times 3$ kernels, 3 batch normalization layers, 3 Leaky ReLU layers, and a $2\times 2$ max-pooling layer. Conv-4 and ResNet-12 generate feature maps of size $19\times 19 \times 64$ and $5\times 5 \times 640$ for $84\times 84$ images, respectively. These feature maps are then mapped through a transformation layer $f_\phi$, which consists of a $1\times 1$ convolutional layer, a batch normalization layer, and a LeakyReLU layer. Finally, $\mathcal{F}_q$ is implemented with two fully connected layers.

\textbf{Training and evaluation details.} During the meta-training phase, we followed the settings in \cite{liu2022dmn4,yan2023few,qiao2024talds}. For Conv-4, we set the learning rate to $1e-3$ and decay $0.1$ every $10$ epoch, training for a total of 30 epochs using the Adam optimizer. For ResNet-12, we pre-trained it first and then conducted meta-training for 40 epochs using momentum SGD with an initial learning rate of $5e-4$ and decay $0.1$ every $10$ epochs.
During the test, as in \cite{liu2022dmn4,yan2023few,qiao2024talds}, we randomly constructed $10000$ episodes from the test set to calculate the classification accuracy. This process was repeated five times, and we reported the average accuracy along with a $95\%$ confidence interval.

\subsection{Comparisons with State-of-the-art Methods}
We choose $7$ generic few-shot learning state-of-the-art baselines\cite{vinyals2016matching,snell2017prototypical,sung2018learning,wertheimer2021few,wang2023meta,kang2023meta}, as well as $5$ SOTA baselines based on LDs\cite{zhang2020deepemd,li2019revisiting,liu2022dmn4,dong2021learning,yan2023few}. For fine-grained datasets, we also selected $9$ state-of-the-art baselines\cite{snell2017prototypical,li2019revisiting,simon2020adaptive,kang2023meta,wang2023meta,zhang2020deepemd,wertheimer2021few,liu2022dmn4,yan2023few}.
\par
\textbf{Results on miniImageNet dataset.}
As shown in Table \ref{tab1}, the performance of our method in the 5-way 1-shot and 5-shot settings exceeds that of all current LDs-based methods \cite{zhang2020deepemd,li2019revisiting,liu2022dmn4,dong2021learning,yan2023few}. Compared to the baseline DN4, our method exhibits significant improvement. In the 5-way 1-shot and 5-shot settings, using Conv-4 as the backbone, it achieved improvements of $5.90\%$ and $4.87\%$, respectively. Compared to the state-of-the-art (SOTA), our method also improved by $1\%$ and $1.21\%$, respectively. When using ResNet-12 as the backbone, improvements of $3.18\%$ and $4.02\%$ were achieved, surpassing SOTA by $1.27\%$ and $0.89\%$, respectively.
\par
\textbf{Results on tieredImageNet dataset.}
As shown in Table \ref{tab1}, our method outperforms the current state-of-the-art LDs-based methods as well. In the 5-way 1-shot and 5-shot settings, when using Conv-4 as the backbone, our method improves by $0.79\%$ and $0.08\%$, respectively, compared to the state-of-the-art method based on LDs. When using ResNet-12 as the backbone, our method improves by $1.14\%$ and $0.89\%$, respectively, compared to the state-of-the-art method based on LDs.
\par
\textbf{Results on fine-grained CUB-200 dataset.}
As shown in Table \ref{tab2}, our method also achieves state-of-the-art performance on fine-grained datasets. In the 5-way 1-shot and 5-shot settings, when using Conv-4 as the backbone, our method improves by $0.26\%$ and $1.68\%$, respectively, compared to the state-of-the-art method based on LDs. When using ResNet-12 as the backbone, our method improves by $1.02\%$ and $1.19\%$, respectively, compared to the state-of-the-art method based on LDs.

\begin{table}[!t]
\caption{
The classification accuracies on the miniImageNet dataset in the 5-way 1-shot and 5-shot settings for backbones with different parameters.}
\label{tab_params}
\small
\resizebox{\linewidth}{!}{
\begin{tabular}{lcccc}
\toprule
\multicolumn{1}{c}{\multirow{2}{*}{\textbf{Method}}} & \multicolumn{1}{c}{\multirow{2}{*}{\textbf{Backbone}}} & \multicolumn{1}{c}{\multirow{2}{*}{$\approx$ \textbf{Params}}} & \multicolumn{2}{c}{\textbf{miniImageNet}}       \\  \cmidrule(r){4-5}
\midrule
CTM\cite{li2019finding}  & ResNet-18  & 11.7 M  & 64.12 $\pm$ 0.82   & 80.51 $\pm$0.13  \\
Neg-Cosine\cite{liu2020negative}  & ResNet-18  & 11.7 M  & 62.33$\pm$0.82  & 80.94 $\pm$0.59  \\
UniSiam+dist\cite{lu2022self}  & ResNet-18  & 11.7 M  & 64.10 ± 0.36  & 82.26$\pm$ 0.25  \\
\midrule
Meta-OLE\cite{kang2023meta} & WRN-28-10  & 36.5 M &75.22$\pm$0.30&86.12$\pm$0.28\\
MetaQDA\cite{zhang2021shallow}  & WRN-28-10  & 36.5 M  & 67.83$\pm$0.64  & 84.28$\pm$0.69  \\
OM\cite{qi2021transductive}  & WRN-28-10  & 36.5 M  & 
66.78$\pm$0.30  & 85.29$\pm$0.41  \\
\midrule
FewTURE\cite{hiller2022rethinking}   & ViT-Small  & 22 M  & 68.02$\pm$\small{0.88}  & 84.51$\pm$\small{0.53}  \\
FewTURE\cite{hiller2022rethinking}   & Swin-Tiny  & 29 M  & \textbf{72.40$\pm$\small{0.78}}  & \textbf{86.38$\pm$\small{0.49}}  \\
\midrule
\textbf{TCDSNet(ours)}  & \multicolumn{1}{c}{ResNet-12} & \multicolumn{1}{c}{\textbf{12.4 M}} & \multicolumn{1}{c}{68.53$\pm$0.19 } & \multicolumn{1}{c}{85.12$\pm$0.42} \\ 
\bottomrule
\end{tabular}
}
\end{table}

\subsection{Ablation Studies}
\textbf{Influence of Top $K$ in support LDs selection.} 
\label{ablation_K}
In subsection \ref{subsec:irbLDs}, we selected $K$ ($K$ as a percentage) LDs based on $\mathcal{CDS}$ for each support class to form a discriminative LDs set. As shown in Table \ref{tab_topk}, we conducted experiments on the miniImagenet and CUB-200 datasets under the 5-way 5-shot setting. When using Conv-4 as the backbone, we set $K$ to $1\%, 2\%, 5\%, 10\%, 30\%$ respectively. When using ResNet-12 as the backbone, we set $K$ to $3\%, 5\%, 10\%, 25\%, 30\%$ respectively. Through experiments, we found that when using Conv-4 as the backbone, the performance is best when $K=2\%$ and $K=10\%$ on both datasets. When using ResNet-12 as the backbone, the performance is best when $K=2\%$ and $K=25\%$ on both datasets. The experimental results indicate that compared to general datasets, fine-grained datasets require more discriminative LDs. Similarly, under the 5-way 1-shot setting, the best performance is achieved with Conv-4 as the backbone when $K=5\%$ and $K=10\%$, and with ResNet-12 as the backbone, the best performance is achieved when $K=5\%$ and $K=10\%$.
\par
\textbf{Comparison with methods using backbones with different parameters.} 
As shown in Table \ref{tab_params}, we selected three baselines\cite{li2019finding,lu2022self,liu2020negative} using ResNet-18 as the backbone, three baselines\cite{kang2023meta,zhang2021shallow,qi2021transductive} using WRN-28-10 as the backbone, and baselines\cite{hiller2022rethinking} using ViT-Small and Swin-Tiny as the backbone. These methods are not LDs-based baselines. Compared to the baselines using ResNet-18 as the backbone, our method outperforms the best-performing method by $4.41\%$ and $2.86\%$ in the 1-shot and 5-shot settings, respectively. When compared to baselines using WRN-28-10 as the backbone\cite{kang2023meta,zhang2021shallow,qi2021transductive}, our method achieves a $0.70\%$ improvement in the 1-shot setting and is only $0.17\%$ lower than OM\cite{qi2021transductive} in the 5-shot setting, despite WRN-28-10 having three times the parameters of ResNet-12. Compared to FewTURE\cite{hiller2022rethinking} with ViT-Small as the backbone, our method achieves improvements of $0.51\%$ and $0.61\%$, and is only $1.26\%$ lower than FewTURE with Swin-Tiny as the backbone in the 5-shot setting. However, Swin-Tiny has 2.3 times the parameters of ResNet-12. Additionally, FewTURE's ViT-Small and Swin-Tiny were trained using 4 and 8 Nvidia A100 40GB GPUs, respectively, making their GPU requirements relatively high.

\section{CONCLUSION}
\label{sec:conclusion}
We propose a novel Task-Aware Contrastive Discriminative Local Descriptor Selection Network (TCDSNet), which utilizes a novel contrastive discriminative measure to filter discriminative local descriptors from the support class. Subsequently, it further selects discriminative query local descriptors from the filtered discriminative support descriptors, ensuring the selection of task-relevant query local descriptors. Extensive experiments validate the superiority and effectiveness of our proposed method. We anticipate that TCDSNet provides a new perspective for research in few-shot learning based on local descriptors.

\section*{Acknowledgment}
This work was supported in part by the National Key R\&D Program of China (2018YFA0701700; 2018YFA0701701) and by the National Natural Science Foundation of China under Grant No.61672364, No.62176172 and No.62002253.

\bibliography{cites.bib}
\bibliographystyle{splncs04}
\end{document}